\documentclass[11pt]{article}

\usepackage[final]{acl}

\usepackage{times}
\usepackage{latexsym}

\usepackage[T1]{fontenc}
\usepackage{inconsolata}

\usepackage[utf8]{inputenc}

\usepackage{microtype}

\usepackage{inconsolata}

\usepackage{graphicx}
\usepackage{url}

\usepackage[utf8]{inputenc}
\usepackage{tikz}
\usetikzlibrary{positioning, arrows.meta, shadows}
\usepackage{xcolor}
\usepackage{inconsolata} 
\definecolor{headerGray}{HTML}{6B7280}
\definecolor{titleDark}{HTML}{111827}
\definecolor{descGray}{HTML}{374151}
\definecolor{boxBorder}{HTML}{E5E7EB}
\definecolor{arrowColor}{HTML}{475569}
\definecolor{boxBlue}{HTML}{F0F7FF} 

\newcommand{\boxcontent}[3]{%
    {\raggedright
    {\color{headerGray}\sffamily\fontsize{7}{8}\selectfont\bfseries #1}\par\vspace{2pt}
    {\color{titleDark}\sffamily\large\bfseries #2}\par\vspace{6pt}
    {\color{descGray}\sffamily\fontsize{8.5}{10}\selectfont #3}\par}
}
\usepackage[most]{tcolorbox}
\usepackage{enumitem}
\usepackage{cleveref}

\definecolor{boxBlue}{HTML}{F0F7FF}
\definecolor{boxBorder}{HTML}{E5E7EB}
\definecolor{titleDark}{HTML}{111827}

\newtcolorbox{promptbox}{
    colback=boxBlue,
    colframe=boxBorder,
    arc=3pt,
    boxrule=0.5pt,
    left=8pt, right=8pt, top=8pt, bottom=8pt,
    fontupper=\small\sffamily, 
    enhanced,
    before skip=10pt,
    after skip=10pt
}
\usepackage{booktabs}
\title{TalkTag}
\title{TalkTag: Automated Fine-Grained Morphosyntactic Error Annotation for Transcribed Speech}
\title{TalkTag:\\Fine-Grained Morphosyntactic Error Annotation for Transcribed Speech}

\author{
  \textbf{Shamira Venturini\textsuperscript{1,2}},
  \textbf{Oliver Hennhöfer\textsuperscript{2}},
  \textbf{Steffen Kinkel\textsuperscript{2}},
  \textbf{Jannik Strötgen\textsuperscript{2}}
\\
\\
  \textsuperscript{1}Karlsruhe Institute of Technology, Germany \\
  \textsuperscript{2}Karlsruhe University of Applied Sciences, Germany
\\
  \small{
   \textbf{Correspondence:} \href{shamira.venturini@h-ka.de}{shamira.venturini@h-ka.de}
  }
}

\begin{document}
\maketitle
\begin{abstract}
Fine-grained morphosyntactic error annotation is important in clinical and developmental language research, yet it is labour-intensive, expert-dependent, and difficult to scale.
We present \texttt{TalkTag}, an LLM-based lightweight tool fine-tuned to automate CHAT-style error annotation in spoken-language transcripts. 
Developed under conditions of extreme data scarcity using children’s narrative data, the system shows the feasibility of linguistic analysis in low-resource settings. 
Our evaluation demonstrates that \texttt{TalkTag} produces encouragingly precise annotation while effectively identifying instances where linguistic ambiguity makes automated tagging genuinely complex.
In summary, with \texttt{TalkTag}, we provide a scalable alternative to manual error annotation and practically viable support for morphosyntactic error annotation.
\end{abstract}

\section{Introduction}

Language Sample Analysis has become an increasingly important method in clinical linguistics and developmental psycholinguistics \citep{MacWhinney2022}. Drawing on naturalistic spoken interaction data, it supports the study of language development and impairment in context. 

During the past decades, TalkBank \footnote{\url{http://talkbank.org}} \citep{macwhinney-etal-2004-talkbank} has substantially advanced the infrastructure for this line of research through large open spoken corpora, the CHAT transcription format, and the CLAN analysis tools \citep{macwhinney2000childes}. CHAT provides a standard way to represent spoken-language transcripts, while CLAN (Computerized Language ANalysis) provides analysis programs for CHAT-formatted files. In this setting, morphosyntactic error codes are written inline on the transcript line, immediately after the form they describe, so that the annotation preserves both the child's production and the analyst's interpretation of the intended target. Although TalkBank includes tools for some aspects of automatic transcription and morphosyntactic annotation, to the best of our knowledge, it currently does not support automatic annotation of morphosyntactic errors.

However, fine-grained morphosyntactic error annotation is important because it provides evidence of grammatical development, impairment, and variation in typically developing \citep{MoraledaSeplveda2022}, as well as populations such as children with developmental language disorders \citep{Leonard2020, Eadie2002, rice-wexler-1996-tense-marker}, Down syndrome \citep{Witecy2023, Katsarou2022, Penke2019}, deaf and hard-of-hearing children with cochlear implant \citep{Benassi2021, Golestani2018}, and autism spectrum disorder \citep{Huang2020}. Manually annotating these phenomena remains slow, labour-intensive, and dependent on expert knowledge. Unlike ordinary morphosyntactic tagging, this task often requires identifying a structured divergence between an erroneous production and an intended target, or recovering morphology that is absent from the surface string but obligatory in context. 
It therefore goes beyond simple sequence labelling, requiring structural linguistic reasoning rather than surface-level pattern matching.

Large Language Models (LLMs) offer a promising solution for this setup by integrating contextual modelling with schema-constrained generation \cite{devlin-etal-2019-bert}. Unlike assigning isolated tags token by token, LLMs can, in principle, evaluate an entire utterance holistically. This allows for the generation of well-formed inline annotations that capture syntactic dependencies and the underlying structural nature of the error.

At the same time, annotated clinical and developmental language data are often scarce, access-restricted, and difficult to use for large-scale model development \citep{AlMarridi2026, Gagliardi2024}. We therefore introduce \texttt{TalkTag}\footnote{\url{http://github.com/OliverHennhoefer/talk-tag}}, a tool for fine-grained morphosyntactic error annotation that follows the CHAT guidelines for word-level error coding. The tool was developed under extremely low-resource conditions, with very limited annotated data \cite{hedderich-etal-2021-survey}. 
To address this, we employed synthetic data augmentation to fine-tune a lightweight, open-weight LLM, effectively expanding the model's exposure to rare error patterns. In this initial prototype, we focus on children’s narrative data, a domain selected not only for its availability but for its high density of developmentally salient morphosyntactic phenomena. These narratives provide a benchmark for evaluating fine-grained annotation capabilities in a complex, real-world linguistic context.

The main contribution of this paper is a lightweight tool for fine-grained morphosyntactic error annotation in clinical and developmental language research, together with its accompanying Python package, \texttt{TalkTag}. More specifically, we formulate CHAT morphosyntactic error coding as a constrained structured-generation task, adapt a small instruction-tuned model to this setting under severe data sparsity, and evaluate the resulting system using automatic scoring, blinded post-hoc adjudication, and human review on unseen corpus material.

The remainder of the paper is structured as follows. Section~\ref{sec:annotation-language} defines the target annotation scheme and the subset of CHAT morphosyntactic error labels modelled in this study. Section~\ref{sec:related-work} then situates the work in relation to prior research on clinical language annotation, error coding, and automatic linguistic analysis. Section~\ref{sec:methods} describes the model, training setup, data, and evaluation design. 
Finally, Section~\ref{sec:results} reports the automatic and human-reviewed results. Section~\ref{sec:discussion} discusses the main linguistic error patterns, the implications of the findings for annotation practice, and the limitations of the current system.

\begin{figure*}[t] 
    \centering
    \resizebox{\textwidth}{!}{
        \begin{tikzpicture}[
            node distance=1.0cm,
            basebox/.style={
                draw=boxBorder,
                fill=boxBlue, 
                rounded corners=6pt,
                drop shadow={opacity=0.08, shadow xshift=1pt, shadow yshift=-1pt},
                inner sep=12pt,
                text width=4.2cm,
                minimum height=3.2cm,
                align=flush left,
                line width=0.5pt
            },
            myarrow/.style={
                -{Stealth[scale=1.2]},
                line width=1.8pt,
                color=arrowColor,
                shorten >=4pt,
                shorten <=4pt
            }
        ]
            \node [basebox] (input) {
                \boxcontent{INPUT}{CHAT input}{Transcript utterance\\plus error scheme}
            };
            
            \node [basebox, right=of input] (adapt) {
                \boxcontent{ADAPT}{TalkTag model}{CHAT-token components\\and LoRA fine-tuning}
            };
            
            \node [basebox, right=of adapt] (generate) {
                \boxcontent{GENERATE}{Candidate tags}{Preserve utterance\\and insert inline tags}
            };
            
            \node [basebox, right=of generate, text width=3.8cm] (use) {
                \boxcontent{USE}{Review}{Automatic scoring\\plus human review}
            };

            \draw [myarrow] (input) -- (adapt);
            \draw [myarrow] (adapt) -- (generate);
            \draw [myarrow] (generate) -- (use);
        \end{tikzpicture}
    }
    \caption{The \texttt{TalkTag} Workflow Pipeline.}
    \label{fig:workflow}
\end{figure*}
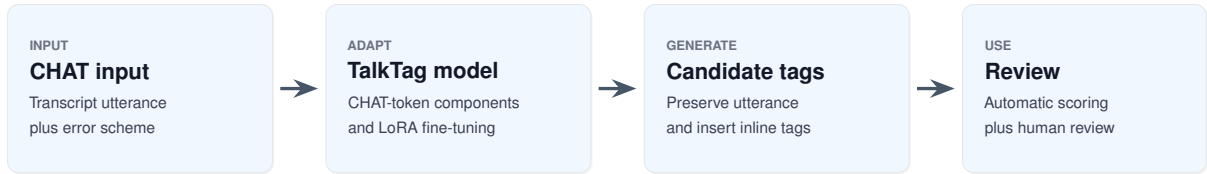

\section{The Annotation Language}
\label{sec:annotation-language}

Within the TalkBank ecosystem, which provides infrastructure for the transcription and analysis of spoken interaction data, the \texttt{MOR} and \texttt{GRASP} programs \citep{macwhinney-2012-morphosyntactic} support automatic morphosyntactic analysis of CHAT transcripts. \texttt{MOR} is a morphological analyser that assigns lexical and grammatical information to each token, producing the \texttt{\%mor} tier with lemma, part-of-speech, and inflectional information. Building on this output, \texttt{GRASP} derives syntactic structure by assigning grammatical relations and dependency-based representations across the utterance on the \texttt{\%gra} tier. Together, these tools make it possible to move from raw transcript text to a linguistically enriched representation of children’s speech. However, their purpose is to recover morphological and syntactic structure rather than to identify, classify, or encode morphosyntactic errors explicitly.

The CHAT Transcription Guidelines provide a general system for marking word-level errors \citep{macwhinney2000childes, chat-manual-part1}. At the level of string form, the annotation tags have a simple and regular surface structure: first, they consist of a fixed bracketed frame following the relevant error, marked with an \texttt{*}.
Next, a flat sequence of colon-separated fields indicates i) the error domain (phonological, semantic, neologistic, dysfluency, and morphological), ii) the error pattern (e.g., missing, superfluous, over-regularised, double-marked morphemes, unknown/known target, etc.) and iii) the morpheme or part-of-speech involved (e.g., past tense, perfective, plural, or third-person singular agreement morphemes; pronouns, prepositions, determiners as parts-of-speech). 
When the intended target is known, it can also be provided in brackets next to the error using the format \texttt{[: target]}. This is used when CLAN's \texttt{MOR} morphological analyser should analyse the target form instead of the produced one, whereas \texttt{[:: target]}\footnote{Since the time of this study, TalkBank has updated the CHAT manual, replacing the \texttt{[:: target]} syntax with \texttt{[= target]}. While the model described here was trained on the earlier convention, the associated Python package includes post-processing to ensure compliance with the latest standard and offers options to toggle the visibility of reconstructions depending on user preference.} can be used to preserve analysis of the produced real-word form while still recording the intended target. 

The surface syntax of the labels is therefore relatively simple, yet their assignment is highly challenging: it often depends on contextual linguistic interpretation and, in many cases, on reconstructing an intended target form.

In this work, we focus on treating [\texttt{* m:*}] labels as mismatches between produced and target forms under lexical identity (i.e., morpheme operations), and [\texttt{* s:r:*}] labels as substitutions involving the same lexical category (e.g., wrong preposition) or [\texttt{* s:r:gc:*}] wrong grammatical category (e.g., adjective for pronoun). Moreover, we use both reconstruction strategies, reserving \texttt{[: target]} for cases where the error produces a non-word form.
Examples of morphological error annotations are:

\begin{itemize}
    \item "\textit{Yesterday I walk} \texttt{[:: walked] [* m:0ed]} \textit{to school}", which marks a missing past tense morpheme. 
    \item "\textit{Yesterday I goed} \texttt{[: went] [* m:=ed]} \textit{to school}", which marks an overregularised past tense morpheme resulting in a non-word form. 
\end{itemize}

Examples of substitutional error annotations are:

\begin{itemize}
    \item "\textit{Yesterday me \texttt{[:: I] [* s:r:gc:pro]} walked} \textit{to school}", which marks a wrong grammatical category of a pronoun. 
    \item "\textit{Yesterday I went in} \texttt{[:: to] [* s:r:prep]} \textit{school}", which marks a wrong preposition. 
\end{itemize}

The CHAT manual provides a standard inventory of error codes, but the framework is extendable: CHAT coding can be adapted to specific applications, and the error-coding system itself allows additional distinctions and combinations within that general format. 
The annotation scheme for morphological and substitutional errors is illustrated in Table \ref{tab:chat-scheme} and Table \ref{tab:chat-scheme2}, respectively. The complete label components inventory is illustrated in Table \ref{tab:chat_scheme_detailed} in Appendix \ref{app:chat-scheme-reference}.

\newcommand{\thickhline}{\noalign{\hrule height 1.1pt}}
\newcommand{\thinhline}{\noalign{\hrule height 0.4pt}}
\begin{table}[h]
\centering
\begin{tabular}{ll}
\thickhline
\textbf{Level 1} & \textbf{Meaning} \\
\thickhline
\texttt{* m:}  & morphosyntactic error \\ 
\thinhline

\thickhline
\textbf{Level 2} & \textbf{Meaning} \\
\thickhline
\texttt{0}       & missing regular form \\ 
\thinhline
\texttt{base:}   & base for irregular form \\ 
\thinhline
\texttt{irr:}    & irregular for base form \\ 
\thinhline
\texttt{sub:}    & past/perfective substitution \\ 
\thinhline
\texttt{=}       & over-regularisation \\ 
\thinhline
\texttt{+}       & superfluous marking\\ 
\thinhline
\texttt{++}      & double marking \\ 
\thinhline
\texttt{vsg:}    & irregular verb 3PS \\ 
\thinhline
\texttt{vun:}    & irregular verb unmarked \\ 
\thinhline
\texttt{allo}    & allomorphic errors \\ 
\thinhline

\thickhline
\textbf{Level 3} & \textbf{Meaning} \\
\thickhline
\texttt{a}       & agreement error \\ 
\thinhline
\texttt{i}       & irregular target \\ 
\thinhline
\texttt{mor}     & target morpheme \\ 
\thinhline
\end{tabular}
\caption{CHAT annotation scheme for morphological errors.}
\label{tab:chat-scheme}
\end{table}

\begin{table}[h]
\centering
\begin{tabular}{ll}
\thickhline
\textbf{Level 1} & \textbf{Meaning} \\
\thickhline
\texttt{* s:}  & substitution error \\ 
\thinhline

\thickhline
\textbf{Level 2} & \textbf{Meaning} \\
\thickhline
\texttt{r:}    & related lexical substitution \\ 
\thinhline
\texttt{r:gc:} & related grammatical substitution \\ 
\thinhline

\thickhline
\textbf{Level 3} & \textbf{Meaning} \\
\thickhline
\texttt{POS} & target morpheme or part-of-speech \\ 
\thinhline
\end{tabular}
\caption{CHAT annotation scheme for substitutional errors.}
\label{tab:chat-scheme2}
\end{table}

\section{Related Work}
\label{sec:related-work}

\paragraph{Grammatical error detection.} Work on grammatical error detection and grammatical error correction addresses linguistic errors more directly, but typically formulates the problem as one of edit detection or sentence-level correction. In this literature, annotation generally starts from a source sentence and a corrected target, from which error spans are identified and labelled with edit operations such as replacement, omission, insertion, or transposition. Schemes such as ERRANT \citep{korre-pavlopoulos-2020-errant} add a further layer of linguistic classification, yielding a structured representation of each edit rather than a flat label. For example, ERRANT combines edit operations such as \texttt{M} (missing), \texttt{R} (replacement), and \texttt{U} (unnecessary) with linguistic error categories to produce fine-grained composite labels. However, the empirical basis of this literature is overwhelmingly written and learner-focused. As noted by \citet{Bryant2023}, the main benchmark datasets are largely derived from L2 English essays, examinations, and learner-platform submissions, including FCE \citep{yannakoudakis-etal-2011-new}, NUCLE \citep{dahlmeier-etal-2013-building}, CoNLL-2013 and 2014 \citep{ng-etal-2013-conll,ng-etal-2014-conll}, Lang-8 \citep{mizumoto-etal-2012-effect,tajiri-etal-2012-tense}, JFLEG \citep{napoles-etal-2017-jfleg}, and W\&I+LOCNESS \citep{bryant-etal-2019-bea}. Consequently, current annotation schemes and correction models are primarily optimized for sentence-level written L2 language, which limits direct transfer to spoken, interactional, or clinically atypical language.

\paragraph{Morphosyntactic error annotation of child language.} More specific work on child language appears limited and idiosyncratic in terms of the annotation scheme and targeted granularity. \citet{Morley2013TheUO} first showed that even relatively coarse linguistic error codes could be sufficient for identifying neurodevelopmental disorders. Building on that result, \citet{morley-etal-2014-data} developed a data-driven dependency-parser approach for detecting and labelling grammatical errors in SALT-annotated transcripts of children’s speech. SALT \citep{miller-etal-2011-salt} supports error coding through a relatively small default inventory of labels, including overgeneralization \texttt{[EO:]}, pronoun error \texttt{[EP:]}, other word-level error \texttt{[EW:]}, extraneous word \texttt{[EW]}, and utterance-level error \texttt{[EU]}, although the active code set can be customised within the software. \citet{Morley2013TheUO} evaluated on the ENNI corpus from CHILDES and the NSR corpus from the SALT database, their system outperformed both Microsoft Word’s grammar checker and a Naive Bayes baseline, while also showing that performance was sensitive to corpus-specific annotation practices and differences in label granularity. This work is therefore highly relevant to the present study, but it remains grounded in a relatively coarse inventory rather than a more fine-grained annotation scheme.

Earlier work by \citet{hassanali-liu-2011-measuring} explored a more fine-grained approach to grammatical error annotation in child language transcripts. Using 677 transcripts from the Paradise corpus \citep{Paradise2005}, they manually annotated ten error categories, with particular attention to verb-related errors, and compared rule-based parse-template methods with statistical classifiers for detecting six error types. Their results showed that statistical approaches generally outperformed rule-based ones. The study’s main contribution was to show that automatic grammar checking could move beyond holistic measures of syntactic development, such as IPSyn \citep{sagae-etal-2005-automatic}, providing a more differentiated profile of grammatical weaknesses. At the same time, it highlighted important limitations, including the difficulty of parsing spoken child language with disfluencies and incomplete utterances, ambiguity in assigning error categories, and the restricted coverage of systems built around a narrow set of constructions.

More recent work has also approached child grammar from other angles: \citet{nikolaus-etal-2024-automatic} developed a context-sensitive scheme for annotating child utterances in caregiver conversation as grammatical, ungrammatical, or ambiguous, and trained Transformer-based models on 4,200 manually annotated CHILDES utterances. Their best models reached near-human agreement and were used to annotate a much larger corpus, confirming that grammaticality increases with age. Unlike work targeting explicit morphosyntactic error labelling, however, their focus was on broad utterance-level grammaticality in a conversational context. 

Most recently, \citet{gebauer25_interspeech} investigated grammatical error detection in spontaneous children’s speech using German kidsTALC data \citep{Rumberg2022}, explicitly addressing both ASR errors and ambiguity in manual error labelling. They proposed a BERT-based recurrent model with iterative pseudo-labelling, showing significant improvements on both manual and automatically transcribed speech. This makes their study particularly relevant to the present work, since it tackles realistic spoken child-language data rather than written text alone. However, the task is still formulated as coarse binary error detection rather than fine-grained morphosyntactic error annotation, so it provides a close methodological precedent without addressing the richer annotation language used in this paper.

Taken together, the literature points to a clear research gap. While grammatical error classification and correction are well developed, this work is largely grounded in written learner-language data and does not transfer straightforwardly to clinical or developmental spoken-language settings. The more specific literature on child spoken-language error analysis is comparatively sparse, and the closest prior systems either date back more than a decade or adopt different annotation scopes and levels of granularity. To our knowledge, there is currently no automated tool for fine-grained, CHAT-compatible morphosyntactic error annotation within the TalkBank ecosystem.

\section{Methods}
\label{sec:methods}
This section describes the study's methodological setup: the model architecture, training regime, data sources, and evaluation strategy for the morphosyntactic error annotation. The workflow pipeline is visually illustrated in Figure \ref{fig:workflow}.

\paragraph{Model.} The model is instruction-tuned base Meta-Llama-3.1-8B-Instruct \citep{llama-3}, loaded in 4-bit quantised form (bnb-4bit) for efficient fine-tuning. Training is implemented using the Unsloth framework, enabling parameter-efficient adaptation via LoRA while keeping the base model weights frozen. This setup allows us to fine-tune an 8B-parameter model under constrained hardware conditions.

The model is instruction-tuned to produce exactly one annotated utterance line, while preserving the original token order, spelling, punctuation, disfluencies, and CHAT symbols. The prompt constrains the model by specifying the structural conditions of the annotation language. The full prompt is provided in Appendix \ref{app:prompt}.

Rather than treating CHAT morphosyntactic labels as a flat inventory of opaque output strings, we treat the annotation scheme as a structured symbolic language. Accordingly, we extend the tokeniser not with full label forms, but with reusable components that recur across the annotation system, including bracketed markers, domain indicators, and subtype fragments. The embedding matrix is resized to accommodate this augmented vocabulary. This reduces fragmentation of CHAT-specific sequences under the base tokeniser and encourages the model to compose licensed tags from meaningful subparts rather than retrieve them from a closed set. This design is motivated by the structure of the annotation scheme itself: CHAT error tags are compositionally organised, encoding contrasts such as domain, operation type, agreement sensitivity, and irregular morphology. The model is thus exposed to the building blocks of a small regular annotation language and must learn to generate well-formed tag combinations under task constraints.

\paragraph{Training.}
Fine-tuning uses LoRA with rank 32, $\alpha=64$, and dropout 0.05. Maximum sequence length is 384 tokens. Training uses a per-device batch size of 8 with gradient accumulation of 4, a warmup ratio of 0.03, and weight decay of 0.1. The model was fine-tuned on an NVIDIA A100-SXM4-80GB GPU. Training took approximately 47 minutes for 3 epochs and 723 optimiser steps.

\paragraph{Data.}
\label{sec:data}

We use CHAT-formatted utterances drawn from a subset of the Edmonton Narrative Norms Instrument (ENNI) corpus \citep{Schneider2006}, available through TalkBank/CHILDES \footnote{\url{https://talkbank.org/childes}} \citep{macwhinney2000childes}. We focus on narratives from children aged 4-5, since this developmental range provides a significant concentration of morphosyntactic phenomena, including overregularisation, agreement errors, tense marking, and clause linking \citep{Cummings2023}. The resulting real-data subset contains 4,585 utterances manually reviewed for the target annotation task. As shown in Table~\ref{tab:data-composition}, this is an extremely low-resource setting \cite{hedderich-etal-2021-survey} not only in corpus size but also in error density, since most real utterances are error-free and many labels have very limited support.  

To mitigate this sparsity, we supplemented the real corpus with curated synthetic examples covering configurations described in the CHAT guidelines. These were generated from error-conditioned prompts and manually reviewed before inclusion. We targeted a minimum overall support of approximately 100 instances per label. We also retained clean utterances in both the real and synthetic portions of the data so that the model would learn when to leave an utterance unchanged and preserve valid CHAT formatting. 

\begin{table}[t]
\centering
\setlength{\tabcolsep}{4pt}
\begin{tabular}{lrrrrr}
\hline
\textbf{Source} & \textbf{Total} & \textbf{0} & \textbf{1} & \textbf{2} & \textbf{$\geq$3} \\
\hline
Synthetic & 5830 & 946  & 4771 & 109 & 4 \\
Real      & 4585 & 4015 & 517  & 47  & 6 \\
\hline
Total     & 10415 & 4961 & 5288 & 156 & 10 \\
\hline
\end{tabular}
\caption{Distribution of utterances by number of annotated errors in the full pre-split master dataset. Columns \textbf{0}, \textbf{1}, \textbf{2}, and \textbf{$\geq$3} indicate the number of utterances containing zero, one, two, or three or more errors, respectively.}
\label{tab:data-composition}
\end{table}

\paragraph{Evaluation.}
We evaluate the system at three complementary levels: automatic scoring on held-out data, followed by a blinded post-hoc review of disagreement cases, and human evaluation of unseen data annotation.

For automatic evaluation, we split both the real ENNI data into train and validation sets as the primary confirmatory benchmarks, as well as augmented data to be used as label-coverage diagnostics. To stabilise rare-label evaluation, we enforced minimum per-label support in the synthetic coverage splits of $N=10$, without distorting the natural error distribution of the real set. This design allows rare-label behaviour to be measured under controlled support while preserving the full real training pool and avoiding aggressive downsampling.

Automatic evaluation combines line-level and label-level perspectives. We report exact match over the full annotated utterance line, but treat it as a secondary summary measure, since the high proportion of clean utterances inflates this score. Our main evaluation metrics focus on the error labels themselves: micro-F1, macro-F1, and per-label precision/recall/F1. Per-label results are further divided into confirmatory and exploratory subsets using minimum-support thresholds, so that claims about individual labels are not based on extremely small counts. Reconstruction targets are reported, but they are not treated as the primary object of evaluation.

Since manual reference annotation may be incomplete or underspecified, automatic scoring can over-penalise outputs that are linguistically plausible but do not match the gold line exactly. We therefore conduct blinded post-hoc adjudication on disagreement cases from the final model. For each reviewed utterance, the reviewer sees the original input and candidate annotations without knowing whether a candidate comes from the gold reference or from the model. Each candidate is assigned one of four labels (\texttt{correct}, \texttt{incorrect}, \texttt{ambiguous}, or \texttt{unsure}) and particularly informative cases are flagged for qualitative analysis. Source identity is stored separately and revealed only after review is complete. These judgments are then merged with the hidden source labels to estimate how often apparent automatic errors reflect genuine model failures, as opposed to ambiguity or omission in the reference annotation. We finally report a conservative updated exact match estimate for the label-bearing subset.

To assess generalisation to unseen data from the same corpus, we run the final model on the remainder of the ENNI corpus, comprising 13,637 utterances not used for training or in-domain evaluation. We manually review all 854 utterances for which the model produced an error annotation, a random sample of 2,200 clean unannotated utterances to estimate the frequency of missed errors, and 91 unannotated but modified utterances. The total reviewed sample is therefore 3,145 utterances. We apply the same coding labels as in the post-hoc review of the automatic evaluation.

\section{Results}
\label{sec:results}
This section reports the results of \texttt{TalkTag} across three complementary stages of evaluation: automatic scoring on held-out training data, blinded post-hoc adjudication, and human review on test inference data. The main question is not whether the system can replace expert annotation, but whether it can provide a reliable first pass that surfaces plausible morphosyntactic error candidates for review.

\subsection{Automatic Model Evaluation}
As shown in Table \ref{tab:final-model-summary}, on the primary test split, the final model achieves 93.6\% exact match, 86.0\% micro-precision, 75.5\% micro-recall, and 80.4\% micro-F1. Validation performance is slightly higher, with 95.4\% exact match and 89.0\% micro-F1. On the synthetic support split, the model reaches 93.2\% micro-F1. Because the real test split is dominated by clean utterances (600/687), we also report results restricted to the 87 utterances containing at least one gold error label. On this subset, the model achieves 66.7\% exact match, 91.4\% precision, 75.5\% recall, and 82.7\% micro-F1. The strongest per-label results on the real test data are obtained for overregularised past morphology (F1 89.4, N = 26), missing third-person singular marking (F1 86.3, N = 25), and allomorphic errors (F1 100.0, N = 14), while pronoun substitution is lower at F1 77.8 (N = 10). This pattern suggests a high-precision annotation aid: proposed labels are usually plausible, but the lower recall indicates that some linguistically valid errors remain difficult to recover automatically, especially when the intended target depends on wider discourse context.

\subsection{Post-hoc Review} Aware that the manual gold annotations may themselves contain errors or omissions, we carried out blinded post-hoc adjudication on all 44 disagreement utterances in the test set: 20 reviewed disagreements were judged acceptable for both model and reference, and in 5 further cases the model output was preferred, yielding a conservative post-hoc acceptable-output rate of 82.8\% (72/87) error-bearing utterances alone.

All the reported results on the real test set are displayed in Table~\ref{tab:final-model-summary}, while synthetic support split (Table \ref{tab:support-results}), label-wise results (Table \ref{tab:appendix_auto_per_label}), and post-hoc details (Table \ref{tab:posthoc-summary}) can be found in the Appendix. 

\begin{table}[t]
\centering
\setlength{\tabcolsep}{4pt}
\begin{tabular}{lrrrrr}
\hline
\textbf{Split} & \textbf{N} & \textbf{EM} & \textbf{P} & \textbf{R} & \textbf{F1} \\
\hline
Val.           & 687 & 95.4\% & 92.0\% & 86.2\% & 89.0\% \\
Test           & 687 & 93.6\% & 86.0\% & 75.5\% & 80.4\% \\
Labels         & 87  & 66.7\% & 91.4\% & 75.5\% & \textbf{82.7}\% \\
Post-hoc       & 87  & \textbf{82.8}\% \\
\hline
\end{tabular}
\caption{Automatic evaluation results. EM denotes full-line exact match; P, R, and F1 are micro-averaged over error tags. Labels restrict evaluation to the 87 tagged utterances in Test; Post-hoc refers to the increased score after manual review of disagreement cases.}
\label{tab:final-model-summary}
\end{table}

\subsection{Human Evaluation of Model Outputs}
Results of the model’s annotation of the unseen portion of the ENNI corpus are summarised in Table~\ref{tab:human_review_summary}. At the utterance level, 93.4\% of reviewed outputs were judged acceptable and 6.6\% incorrect. At the label level, 83.7\% of reviewed label-bearing cases were judged correct. Among the 146 unaccepted label judgments, 94 were false positives, 31 were incorrect labels, and 20 were false negatives. In the audit of 2,200 model-clean utterances, 7 probable missed errors were identified.

At the label level, agreement-related labels were the most frequent in the reviewed predictions. Grouping the three main agreement labels -- missing third-person singular marking (e.g., \textit{he go} for \textit{goes}), irregular unmarked for singular (e.g., \textit{he are/were} for \textit{is/was}), and irregular singular for unmarked (e.g., \textit{they is/was} for \textit{are/were}) -- yields 408 reviewed instances in total. The next most frequent labels were allomorphic errors (105), overregularised past forms (84), and pronoun substitutions (70). Among past-related labels, the most frequent pattern was substitution of the base form for an irregular past form (e.g., \textit{go} for \textit{went}). Accuracy for these frequent labels was 87.3\% for the agreement group as a whole, 92.4\% for allomorphic errors, 78.6\% for overregularised past forms, 71.4\% for pronoun substitutions, and 82.1\% for base-for-irregular past substitutions. Detailed per-label results are reported in Table~\ref{tab:appendix_enni_per_label} in the Appendix.

\begin{table}[t]
\centering
\small
\begin{tabular}{lrr}
\hline
\textbf{Measure} & \textbf{Count} & \textbf{Percent} \\
\hline
Reviewed utterances (raw) & 3145 & 100.0\% \\
Out-of-scope exclusions & 22 & 0.7\% \\
Official reviewed total & 3123 & -- \\
\hline
Utterance-level acceptable & 2917 & 93.4\% \\
Utterance-level incorrect & 206 & 6.6\% \\
\hline
Label-level reviewed cases & 894 & -- \\
Label-level correct & 748 & 83.7\% \\
Label-level incorrect & 146 & 16.3\% \\
\quad Incorrect label & 31 & 3.5\% \\
\quad False negative & 20 & 2.2\% \\
\quad False positive & 94 & 10.5\% \\
\hline
Audited model-clean utterances & 2200 & -- \\
Probable missed errors & 7 & 0.32\% \\
\hline
\end{tabular}
\caption{Human-reviewed evaluation on unseen ENNI data. Percentages for utterance-level outcomes are computed over the reviewed total ($N=3123$). Label-level percentages are computed over reviewed label-bearing cases ($N=894$). The clean-audit miss rate is computed over the audited model-clean sample ($N=2200$).}
\label{tab:human_review_summary}
\end{table}

Among the qualitatively reviewed cases, the clearest and most recurrent linguistic patterns causing confusion involved uninflected verb forms, especially the interaction between tense and agreement and cases of invariant verbs (irregular verbs whose past form is identical to the base form). We focus on this pattern here because it directly addresses a marker of developmental language-disordered speech: preference for uninflected verb forms.

We found 16/17 cases in which the model erroneously assigned the agreement label to uninflected verbs following a third-person singular subject, even though an obligatory context for licensing the past tense was present. Of these, 8 cases involved an invariant verb. We found five additional cases in which invariant verbs were overtly overregularised (e.g., \textit{hurted}, \textit{costed}, \textit{putted}); in these cases, the model did not converge on a single analysis but alternated between overregularisation and double-marking annotation.

\section{Discussion}
\label{sec:discussion}

Taken together, the results suggest that a useful tool for fine-grained morphosyntactic error annotation can be developed even under conditions of extreme data sparsity and limited computational resources. Although \texttt{TalkTag}'s scores on the full test set are partly inflated by the large number of error-free utterances, performance remains encouraging on genuinely error-bearing cases. In addition, the blinded post-hoc review shows that a non-trivial subset of apparent disagreements reflects linguistic ambiguity or underspecification in the reference annotation rather than straightforward model failure. This is important both for metric interpretation and for practical use: full-line disagreement might not always correspond to a linguistically unacceptable output.

\paragraph{Clinical and developmental relevance.} The results are especially encouraging for agreement-related errors, overregularised past forms and pronoun substitutions, which were among the most frequent and best-supported categories in the reviewed data. These categories are also linguistically meaningful. Tense and agreement morphology, as well as difficulties with pronouns, are central to the study of developmental language disorder and autism spectrum disorder \cite{Leonard2020, Eadie2002, rice-wexler-hershberger-1998-tense-over-time, wexler-schutze-rice-1998-subject-case, rice-wexler-1996-tense-marker}. Overregularisation is a well-established feature of typical language development and remains informative when it persists or occurs at elevated rates in atypical development \citep{MoraledaSeplveda2022, marcus-etal-1992-overregularization}. From this perspective, the model’s relative success on these labels is encouraging not only in engineering terms but also because it aligns with clinically and developmentally relevant dimensions of child language.

\paragraph{Agreement bias in tense-agreement ambiguities.} At the same time, the qualitative review revealed a clear and recurrent failure mode involving bare verb forms following third-person singular subjects. In these cases, the model often preferred agreement-based analyses over missing past-tense interpretations, including in some contexts where the surrounding discourse licensed a past reading. This pattern was especially pronounced for zero-change verbs such as \textit{hurt} and \textit{put}, whose past forms are identical to their base forms. More generally, zero-change verbs formed a persistent challenge in the reviewed sample, and when they were overtly overregularised (e.g., \textit{hurted}, \textit{costed}, \textit{putted}), the model alternated between overregularisation and double-marking analyses rather than converging on a single label decision.

These patterns are informative because they suggest that the model’s errors are not simply random, but partly structured by a preference for locally recoverable agreement analyses over broader tense interpretation. One plausible explanation is that agreement errors are both more locally diagnosable and more strongly represented in the training signal, whereas missing past-tense interpretations often require integration of wider temporal and discourse context. At the same time, these cases also highlight a genuine property of the task itself: in spontaneous non-standard speech, morphosyntactic annotation is often difficult precisely because sparse morphology, discourse context, and lexical irregularity interact.

\paragraph{Generalisation and pre-annotation utility.} The broader review on unseen ENNI material seems to reinforce this picture. Although overall acceptability remained high, the error profile on this larger sample differed somewhat from that of the held-out test set: agreement-related labels again dominated the reviewed predictions, but the main source of degradation was over-annotation rather than omission. The audit of model-clean utterances nevertheless suggests that silent misses remain comparatively infrequent. This pattern supports a cautious interpretation of the tool as one that is more useful for surfacing plausible candidate errors than for producing final annotations without review.

From a practical perspective, the present results suggest that the tool is already useful as a pre-annotation aid, even where its outputs still require human correction. Prior work on machine-assisted annotation has shown that automatic pre-annotation can reduce annotation effort, improve consistency, and increase annotation speed without harming final quality \citep{Lingren2014,fort-sagot-2010-influence,mikulova-etal-2022-quality}. The contribution of the present system is therefore not that it eliminates the need for expert review, but that it provides a linguistically informed first-pass annotation over a large volume of CHAT material. This is especially valuable in a domain where fully manual annotation is slow, costly, and itself subject to ambiguity and inconsistency.

\paragraph{Scope of the present prototype.} At the same time, the present findings should be interpreted within the scope of the current prototype. The model was developed and evaluated on children’s narrative data from a single corpus family, and the synthetic support split serves only as secondary evidence of label-space coverage under controlled conditions rather than as a substitute for naturalistic evaluation. Moreover, some reviewed errors were tied to pre-existing CHAT annotations or discourse-formatting cues rather than to morphosyntactic analysis alone.

\section*{Limitations}
\label{sec:limitations}

The present study should be understood as a prototype rather than a complete solution to CHAT-style error annotation. The model was developed under conditions of extreme data scarcity and trained on a reduced subset of the CHAT error inventory, focusing on selected morphosyntactic and closely related substitution labels. As a result, the current system does not yet cover the full range of CHAT-compatible error phenomena, and support for some rare or irregular patterns remains limited. This is particularly relevant for error types that were only sparsely represented in the available training data. Importantly, this is not simply an artefact of the present experiment: some of these phenomena are genuinely infrequent in naturalistic corpora, which makes it inherently difficult to obtain enough real examples for robust learning.

The current annotation scope is also intentionally narrow. For example, forms such as \textit{they going to the shop} may plausibly be analysed not as cases of superfluous progressive marking, but as instances of missing auxiliary. However, the present model was not trained to annotate missing parts of speech, and such cases therefore fall outside the annotation scope of the current system. Within that restricted label space, we treated the model’s superfluous-progressive analysis as acceptable, since it captures a genuine deviation while avoiding unsupported labels. More generally, the prototype is better understood as a first step toward fine-grained morphosyntactic error annotation than as a holistic grammar annotation tool.

These constraints also make it important to understand what the model is learning under conditions of scarcity. The present study does not disentangle the mechanisms by which the model produces its annotations. The fine-tuning was designed to encourage recovery of an intended target form while generating the final inline label, making the approach loosely related to an analysis-by-synthesis perspective. At the same time, the model was also allowed to generate CHAT labels compositionally rather than retrieve them from a fixed inventory of whole forms. The current results, therefore, do not establish whether the observed gains arise from implicit target-form reconstruction, from better formatting control, or from a learned mapping between linguistic error patterns and the internal structure of the annotation language. Future work will investigate these possibilities more directly, including whether improvements are concentrated in error types that require target-form recovery and whether performance depends on latent correction-like inference.

A further limitation concerns generalisation. Although the model was evaluated both on held-out data and on new unseen material, all real-data evaluation was conducted within the ENNI corpus. This provides a useful first test of robustness, but it does not establish how well the model transfers across other CHILDES corpora, elicitation settings, age ranges, or clinical populations. Future work should therefore test the system on additional child-language corpora, as well as on more clearly out-of-distribution material such as second-language learner data and adult's clinical speech.

These limitations point to the main conditions under which the prototype should be used: as a human-in-the-loop pre-annotation aid rather than as a fully automatic replacement for expert judgement. Broader annotation coverage, greater robustness to authentic CHAT markup, and wider evaluation across corpora and populations will be needed before the system can be treated as a more general annotation tool.

\section*{Acknowledgments}
This work was supported by the Ministry of Science, Research and the Arts of Baden-Württemberg (MWK) and KIT's Accessibility through AI-based Assistive Technology (KATE) Graduate School.

\paragraph{Generative AI Assistance Declaration} During the preparation of this work, the author(s) used ChatGPT to rephrase, proofread or summarise text content. After using this tool, the author(s) reviewed and edited the content as needed and take(s) full responsibility for the publication’s content.

\paragraph{Data Availability Statement} The data used in this study are not redistributed with our code release. They are hosted by TalkBank/CHILDES and should be obtained directly from the official source under the applicable TalkBank access and licensing rules.

\bibliography{custom2}

\appendix
\clearpage

\begin{figure*}[!htbp] 
\section{Annotation Prompt}
\label{app:prompt}
\begin{promptbox}
    \textbf{\color{titleDark}Role.} You are a TalkBank CHAT annotator for morphosyntactic error coding.
    \smallskip
    
    \textbf{\color{titleDark}Task.} Annotate the input utterance by inserting valid CHAT error tags inline.
    \smallskip
    
    \textbf{\color{titleDark}Output requirements.}
    \begin{enumerate}[label=\arabic*., nosep, leftmargin=1.5em, itemsep=4pt, topsep=4pt]
        \item Preserve original token order, spelling, casing, punctuation, disfluencies, and CHAT symbols.
        \item Do not rewrite, paraphrase, or correct the utterance.
        \item Insert only error tags inline, following the error token.
        \item If no target error is present, return the utterance unchanged.
        \item Write the correct target form as \texttt{[: target]} when the incorrect morpheme yields a nonword, and as \texttt{[:: target]} when the error is an attested word.
        \item Build each CHAT error tag compositionally from licensed scheme parts rather than relying on a memorised whole-label form.
        \item Use \texttt{m:*} only for same-lexeme morphological contrasts and \texttt{s:*} only for substitutional contrasts.
        \item Use \texttt{:a} only for agreement-sensitive labels that license it.
        \item Use \texttt{:i} only where an irregular-sensitive label licenses it.
        \item Output only licensed CHAT tags; do not invent unattested or unsupported combinations.
        \item Output exactly one annotated utterance line and nothing else.
    \end{enumerate}
\end{promptbox}
\end{figure*}

\begin{table*}[t]
\section{Training Label Inventory}
\label{app:train-labels-inv}
\centering
\footnotesize
\setlength{\tabcolsep}{10pt}
\renewcommand{\arraystretch}{1.04}
\begin{tabular}{llll}
\hline
\textbf{Label} & \textbf{Label} & \textbf{Label} & \textbf{Label} \\
\hline
\texttt{[* m:++ed:i]} & \texttt{[* m:+ing]} & \texttt{[* m:base:er]} & \texttt{[* m:vsg:a]} \\
\texttt{[* m:++ed]} & \texttt{[* m:+s:a]} & \texttt{[* m:base:est]} & \texttt{[* m:vun:a]} \\
\texttt{[* m:++en:i]} & \texttt{[* m:+s]} & \texttt{[* m:base:s]} & \texttt{[* s:r:der]} \\
\texttt{[* m:++s:i]} & \texttt{[* m:0's]} & \texttt{[* m:irr:ed]} & \texttt{[* s:r:gc:det]} \\
\texttt{[* m:++s]} & \texttt{[* m:03s:a]} & \texttt{[* m:irr:en]} & \texttt{[* s:r:gc:pro]} \\
\texttt{[* m:+3s:a]} & \texttt{[* m:0ed]} & \texttt{[* m:irr:s]} & \texttt{[* s:r:prep]} \\
\texttt{[* m:+3s]} & \texttt{[* m:0ing]} & \texttt{[* m:sub:ed]} & \\
\texttt{[* m:+ed:i]} & \texttt{[* m:0s:a]} & \texttt{[* m:sub:en]} & \\
\texttt{[* m:+ed]} & \texttt{[* m:=ed]} & \texttt{[* m:base:ed]} & \\
\texttt{[* m:+en]} & \texttt{[* m:=en]} & \texttt{[* m:base:en]} & \\
\texttt{[* m:=s]} & \texttt{[* m:allo]} &  & \\
\hline
\end{tabular}
\caption{Full training-set label inventory for the final confirmatory model package. The table lists all CHAT morphosyntactic error labels seen during training, irrespective of whether later evaluation support for a label comes from real or synthetic data.}
\label{tab:appendix_training_label_inventory}
\end{table*}

\begin{table*}[t]
\section{Additional Evaluation Results}
\label{app:additional-results}
\centering
\footnotesize
\setlength{\tabcolsep}{5pt}
\begin{tabular}{lrrrrr}
\hline
\textbf{Split} & \textbf{N} & \textbf{EM} & \textbf{P} & \textbf{R} & \textbf{F1} \\
\hline
Val.   & 370 & 82.4\% & 91.9\% & 88.7\% & 90.3\% \\
Test   & 370 & 82.7\% & 94.5\% & 92.0\% & 93.2\% \\
\hline
\end{tabular}
\caption{Automatic evaluation on the synthetic support splits. Exact denotes full-line exact match against the gold annotated utterance; P, R, and F1 are micro-averaged over CHAT error tags.}
\label{tab:support-results}
\end{table*}

\begin{table*}[t]
\centering
\footnotesize
\setlength{\tabcolsep}{7pt}
\begin{tabular}{lrrrrrr}
\hline
\textbf{Subset} & \textbf{N} & \textbf{Prev. exact} & \textbf{Reviewed} & \textbf{Both acc.} & \textbf{Model pref.} & \textbf{Post-hoc acc.} \\
\hline
Test real (full)     & 687 & 643 (93.6\%) & 44 & 20 & 5 & 97.2\% \\
Test real (labelled) & 87  & 58 (66.7\%)  & 29 & 11 & 3 & 82.8\% \\
\hline
\end{tabular}
\caption{Post-hoc adjudication results on \texttt{test\_real}. Previous exact gives the number of exact automatic matches before manual review. Post-hoc acceptable counts exact matches plus reviewed disagreement cases judged acceptable for both outputs or preferred for the model.}
\label{tab:posthoc-summary}
\end{table*}

\begin{table*}[t]
\centering
\footnotesize
\setlength{\tabcolsep}{5pt}
\renewcommand{\arraystretch}{0.95}
\begin{tabular}{lrrrr@{\qquad}lrrrr}
\hline
\multicolumn{5}{l}{\textbf{test\_real}} & \multicolumn{5}{l}{\textbf{test\_coverage}} \\
\textbf{Label} & \textbf{P} & \textbf{R} & \textbf{F1} & \textbf{N} & \textbf{Label} & \textbf{P} & \textbf{R} & \textbf{F1} & \textbf{N} \\
\hline
\texttt{[* m:allo]}     & 100.0 & 100.0 & 100.0 & 14 & \texttt{[* m:=en]}      & 100.0 & 100.0 & 100.0 & 12 \\
\texttt{[* m:vun:a]}    & 100.0 & 100.0 & 100.0 & 2  & \texttt{[* m:irr:en]}   & 100.0 & 100.0 & 100.0 & 11 \\
\texttt{[* s:r:der]}    & 100.0 & 100.0 & 100.0 & 1  & \texttt{[* m:=ed]}      & 100.0 & 100.0 & 100.0 & 10 \\
\texttt{[* m:=ed]}      & 100.0 & 80.8  & 89.4  & 26 & \texttt{[* m:++en:i]}   & 100.0 & 100.0 & 100.0 & 10 \\
\texttt{[* m:03s:a]}    & 84.6  & 88.0  & 86.3  & 25 & \texttt{[* m:+en]}      & 100.0 & 100.0 & 100.0 & 10 \\
\texttt{[* s:r:gc:pro]} & 87.5  & 70.0  & 77.8  & 10 & \texttt{[* m:0's]}      & 100.0 & 100.0 & 100.0 & 10 \\
\texttt{[* m:base:ed]}  & 75.0  & 75.0  & 75.0  & 4  & \texttt{[* m:base:en]}  & 100.0 & 100.0 & 100.0 & 10 \\
\texttt{[* m:++ed:i]}   & 100.0 & 50.0  & 66.7  & 2  & \texttt{[* m:base:er]}  & 100.0 & 100.0 & 100.0 & 10 \\
\texttt{[* m:++ed]}     & 50.0  & 100.0 & 66.7  & 1  & \texttt{[* m:base:est]} & 100.0 & 100.0 & 100.0 & 10 \\
\texttt{[* m:vsg:a]}    & 50.0  & 50.0  & 50.0  & 2  & \texttt{[* m:irr:ed]}   & 100.0 & 100.0 & 100.0 & 10 \\
\texttt{[* s:r:prep]}   & 50.0  & 33.3  & 40.0  & 3  & \texttt{[* m:sub:ed]}   & 100.0 & 100.0 & 100.0 & 10 \\
\texttt{[* m:+ed]}      & 0.0   & 0.0   & 0.0   & 2  & \texttt{[* m:sub:en]}   & 100.0 & 100.0 & 100.0 & 10 \\
\texttt{[* m:0ing]}     & 0.0   & 0.0   & 0.0   & 2  & \texttt{[* m:++s]}      & 90.9  & 100.0 & 95.2  & 10 \\
\texttt{[* s:r:gc:det]} & 0.0   & 0.0   & 0.0   & 2  & \texttt{[* m:+s:a]}     & 90.9  & 100.0 & 95.2  & 10 \\
\texttt{[* m:0ed]}      & 0.0   & 0.0   & 0.0   & 1  & \texttt{[* m:irr:s]}    & 90.9  & 100.0 & 95.2  & 10 \\
\texttt{[* m:0s:a]}     & 0.0   & 0.0   & 0.0   & 1  & \texttt{[* m:+ing]}     & 90.0  & 90.0  & 90.0  & 10 \\
                         &       &       &       &    & \texttt{[* m:+3s]}      & 81.2  & 100.0 & 89.7  & 13 \\
                         &       &       &       &    & \texttt{[* m:=s]}       & 100.0 & 80.0  & 88.9  & 10 \\
                         &       &       &       &    & \texttt{[* m:base:s]}   & 100.0 & 80.0  & 88.9  & 10 \\
                         &       &       &       &    & \texttt{[* m:++s:i]}    & 88.9  & 80.0  & 84.2  & 10 \\
                         &       &       &       &    & \texttt{[* m:+3s:a]}    & 100.0 & 60.0  & 75.0  & 10 \\
                         &       &       &       &    & \texttt{[* m:03s:a]}    & 80.0  & 66.7  & 72.7  & 12 \\
\hline
\end{tabular}
\caption{Per-label automatic evaluation grouped by evaluation source. Values are percentages except for support (N). Real rows are taken from \texttt{test\_real}; synthetic rows are labels absent from \texttt{test\_real} and therefore reported on \texttt{test\_coverage}.}
\label{tab:appendix_auto_per_label}
\end{table*}

\begin{table*}[t]
\centering
\footnotesize
\setlength{\tabcolsep}{8pt}
\renewcommand{\arraystretch}{0.96}
\begin{tabular}{lrr@{\qquad}lrr}
\hline
\textbf{Label} & \textbf{Correct} & \textbf{N} & \textbf{Label} & \textbf{Correct} & \textbf{N} \\
\hline
\texttt{[* m:03s:a]}    & 88.7\%  & 328 & \texttt{[* m:sub:en]}   & 100.0\% & 6 \\
\texttt{[* m:allo]}     & 92.4\%  & 105 & \texttt{[* s:r:prep]}   & 83.3\%  & 6 \\
\texttt{[* m:=ed]}      & 78.6\%  & 84  & \texttt{[* m:+ed]}      & 100.0\% & 5 \\
\texttt{[* s:r:gc:pro]} & 71.4\%  & 70  & \texttt{[* m:0s:a]}     & 75.0\%  & 4 \\
\texttt{[* m:vun:a]}    & 68.2\%  & 44  & \texttt{[* m:0's]}      & 50.0\%  & 4 \\
\texttt{[* m:vsg:a]}    & 97.2\%  & 36  & \texttt{[* m:irr:s]}    & 50.0\%  & 4 \\
\texttt{[* m:base:ed]}  & 82.1\%  & 28  & \texttt{[* m:=en]}      & 33.3\%  & 3 \\
\texttt{[* s:r:der]}    & 23.8\%  & 21  & \texttt{[* m:++s:i]}    & 50.0\%  & 2 \\
\texttt{[* m:irr:ed]}   & 86.7\%  & 15  & \texttt{[* m:+3s:a]}    & 50.0\%  & 2 \\
\texttt{[* m:++ed:i]}   & 76.9\%  & 13  & \texttt{[* m:++est]}    & 100.0\% & 1 \\
\texttt{[* m:++ed]}     & 61.5\%  & 13  & \texttt{[* m:+en]}      & 100.0\% & 1 \\
\texttt{[* m:0ing]}     & 100.0\% & 11  & \texttt{[* m:base:en]}  & 100.0\% & 1 \\
\texttt{[* m:+3s]}      & 54.5\%  & 11  & \texttt{[* m:irr:en]}   & 100.0\% & 1 \\
\texttt{[* m:0ed]}      & 80.0\%  & 10  & \texttt{[* m:=ed:i]}    & 0.0\%   & 1 \\
\texttt{[* s:r:gc:det]} & 50.0\%  & 10  & \texttt{[* m:=ing]}     & 0.0\%   & 1 \\
\texttt{[* m:sub:ed]}   & 100.0\% & 9   & \texttt{[* m:base:der]} & 0.0\%   & 1 \\
\texttt{[* m:+ing]}     & 100.0\% & 8   &                          &          &   \\
\hline
\end{tabular}
\caption{Per-label human-reviewed exactness on the rest of the ENNI corpus, sorted by support. Because ENNI does not have exhaustive gold annotation, this table does not report recall or F1.}
\label{tab:appendix_enni_per_label}
\end{table*}

\begin{table*}[t]
\section{Expanded CHAT Scheme Reference}
\label{app:chat-scheme-reference}
\centering
\footnotesize
\setlength{\tabcolsep}{4pt} 
\renewcommand{\arraystretch}{1.1}
\begin{tabular}{ll p{0.5cm} ll p{0.5cm} ll}
\hline
\multicolumn{2}{c}{\textbf{Level 1}} & & \multicolumn{2}{c}{\textbf{Level 2}} & & \multicolumn{2}{c}{\textbf{Level 3}} \\
\cline{1-2} \cline{4-5} \cline{7-8}
\textbf{Code} & \textbf{Meaning} & & \textbf{Code} & \textbf{Meaning} & & \textbf{Code} & \textbf{Meaning} \\
\hline
\texttt{[* m:]}  & morphosyntactic error & & \texttt{0}      & missing              & & \texttt{-ing} & progressive \\
\texttt{[* s:]}  & substitution error    & & \texttt{base:} & bare form            & & \texttt{-3s}  & 3SG         \\
                 &                       & & \texttt{sub:}  & substitution         & & \texttt{-ed}  & past        \\
                 &                       & & \texttt{irr:}  & irregular            & & \texttt{-en}  & perfective  \\
                 &                       & & \texttt{=}     & over-regularisation  & & \texttt{-s}   & plural      \\
                 &                       & & \texttt{+}     & superfluous          & & \texttt{'s}   & possessive  \\
                 &                       & & \texttt{++}    & double marking       & & \texttt{-s'}  & plural possessive\\
                 &                       & & \texttt{vsg:}  & irregular verb 3SG   & & \texttt{-er}  & comparative \\
                 &                       & & \texttt{vun}   & irregular verb unmarked   & & \texttt{-est} & superlative \\
                 &                       & & \texttt{allo}  & allomorph            & & \texttt{a}    & agreement   \\
                 &                       & & \texttt{s:r:}  & related lexical substitution    & & \texttt{i}    & irregular   \\
                 &                       & & \texttt{s:r:gc:}& related grammatical substitution    & & \texttt{POS}  & target POS  \\
\hline
\end{tabular}
\caption{Expanded reference table for CHAT-style error-label components used in this study.}
\label{tab:chat_scheme_detailed}
\end{table*}

\end{document}